%% file: acl_latex.tex
\pdfoutput=1

\documentclass[11pt]{article}

\usepackage[preprint]{acl}

\usepackage{times}
\usepackage{latexsym}

\usepackage[T1]{fontenc}

\usepackage[utf8]{inputenc}

\usepackage{microtype}

\usepackage{inconsolata}

\usepackage{graphicx}

\usepackage[utf8]{inputenc} 
\usepackage[T1]{fontenc}    
\usepackage{hyperref}       
\usepackage{url}            
\usepackage{booktabs}       
\usepackage{amsfonts}       
\usepackage{nicefrac}       
\usepackage{microtype}      
\usepackage{xcolor}         
\usepackage{multirow,multicol}
\usepackage{graphicx}
\usepackage{arydshln}
\usepackage{amsmath}
\usepackage{enumitem}
\usepackage{siunitx}
\usepackage{mdframed}
\usepackage{epigraph}
\usepackage{listings}
\usepackage{nicematrix} 
\usepackage[most]{tcolorbox}
\usepackage[table]{xcolor}
\usepackage{tabularx}
\usepackage{colortbl}
\usepackage{caption}

\definecolor{groupgray}{gray}{0.95}
\definecolor{mygreen}{HTML}{148F77}
\definecolor{myred}{HTML}{E74C3C}
\definecolor{myblue}{HTML}{3498DB}

\definecolor{codegreen}{rgb}{0,0.6,0}
\definecolor{codegray}{rgb}{0.5,0.5,0.5}
\definecolor{codepurple}{rgb}{0.58,0,0.82}
\definecolor{backcolour}{rgb}{0.96,0.96,0.98}
\definecolor{denim}{rgb}{0.08, 0.38, 0.74}
\definecolor{customGreen}{HTML}{2AA587} 
\definecolor{customBlue}{HTML}{377EB8}  
\definecolor{customPurple}{HTML}{7570b3}
\definecolor{gptColor}{HTML}{388E3C}
\definecolor{qwenColor}{HTML}{9575CD}

\lstdefinestyle{mystyle}{
    backgroundcolor=\color{backcolour},   
    commentstyle=\color{codegreen},
    keywordstyle=\color{magenta},
    numberstyle=\tiny\color{codegray},
    stringstyle=\color{codepurple},
    basicstyle=\ttfamily\small,
    breakatwhitespace=false,         
    breaklines=true,                 
    captionpos=b,                    
    keepspaces=true,                 
    numbers=left,                    
    numbersep=5pt,                  
    showspaces=false,                
    showstringspaces=false,
    showtabs=false,                  
    tabsize=2,
    frame=none
}
\title{Codified Foreshadowing-Payoff Text Generation}

\author{Longfei Yun, Kun Zhou, Yupeng Hou, Letian Peng\thanks{Corresponding author.}, Jingbo Shang \\
University of California, San Diego \\
  \texttt{\{loyun, kuzhou, yphou, lepeng, jshang\}@ucsd.edu}
}

\usepackage{xspace}

\begin{document}
\maketitle
\begin{abstract}
    \input{0-abs}
\end{abstract}

\input{1-intro}
\input{2-rel}
\input{3-method}

\input{4-setting}
\input{5-exp}
\input{6-con}
\input{7-lim}

\clearpage
\bibliography{custom}

\clearpage

\appendix
\input{appendix}

\end{document}

%% file: 0-abs.tex
Foreshadowing and payoff are ubiquitous narrative devices through which authors introduce commitments early in a story and resolve them through concrete, observable outcomes. However, despite advances in story generation, large language models (LLMs) frequently fail to bridge these long-range narrative dependencies, often leaving "Chekhov's guns" unfired even when the necessary context is present. Existing evaluations largely overlook this structural failure, focusing on surface-level coherence rather than the logical fulfillment of narrative setups.

In this paper, we introduce Codified Foreshadowing-Payoff Generation (CFPG), a novel framework that reframes narrative quality through the lens of payoff realization. Recognizing that LLMs struggle to intuitively grasp the "triggering mechanism" of a foreshadowed event, CFPG transforms narrative continuity into a set of executable causal predicates. By mining and encoding Foreshadow-Trigger-Payoff triples from the BookSum corpus, we provide structured supervision that ensures foreshadowed commitments are not only mentioned but also temporally and logically fulfilled.

Experiments demonstrate that CFPG significantly outperforms standard prompting baselines in payoff accuracy and narrative alignment. Our findings suggest that explicitly codifying narrative mechanics is essential for moving LLMs from surface-level fluency to genuine narrative competence.\footnote{Code:  \href{https://github.com/LongfeiYun17/CFPG}{https://github.com/LongfeiYun17/CFPG}}

%% file: 1-intro.tex
\section{Introduction}

\epigraph{\textit{``If in the first act you have hung a pistol on the wall, then in the following one it should be fired.''}}
{\textemdash \textit{Anton Chekhov}}

Human-authored narratives commonly rely on foreshadowing and payoff to establish coherence over the course of a story~\citep{riedl2010narrative, prince2003dictionary}. By explicitly introducing objects, intentions, or conditions early on, authors create narrative commitments that are expected to be concretely resolved later~\citep{Todorov1969StructuralAO}. The successful realization of these commitments, rather than mere sentence-level fluency, is a defining property of competent storytelling~\citep{forster1956aspects}.

\begin{figure}[t] 
    \centering
     \includegraphics[width=0.49\textwidth]{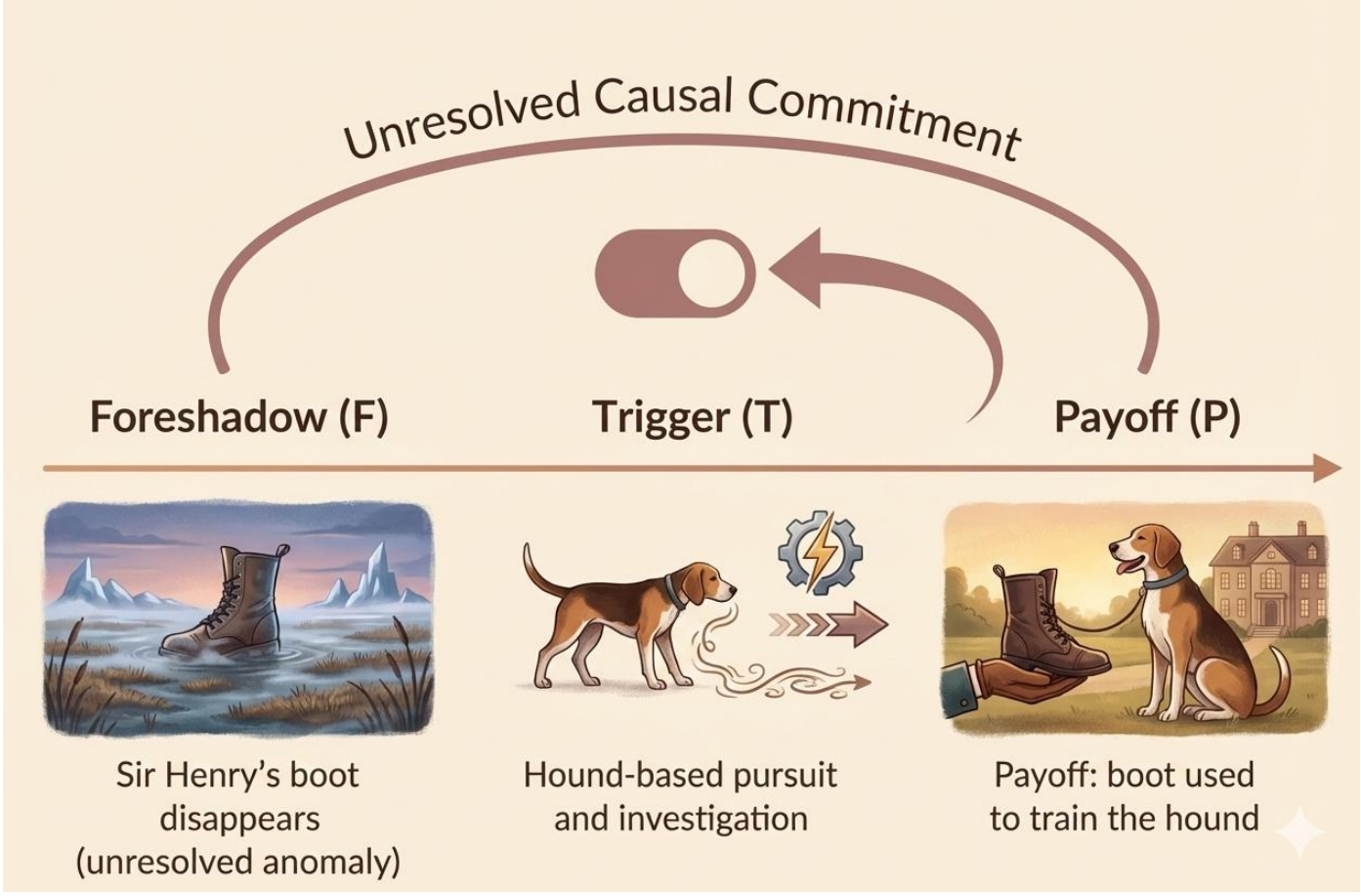}
    \caption{Illustration of Foreshadow–Trigger–Payoff decomposition using a narrative example from \textit{The Hound of the Baskervilles}. The disappearance of the boot introduces an unresolved causal commitment, which remains dormant until a triggering narrative condition activates its resolution.
}
    \label{fig:problem_formulation}
\end{figure}

\begin{figure*}[htbp] 
    \centering
     \includegraphics[width=0.98\textwidth]{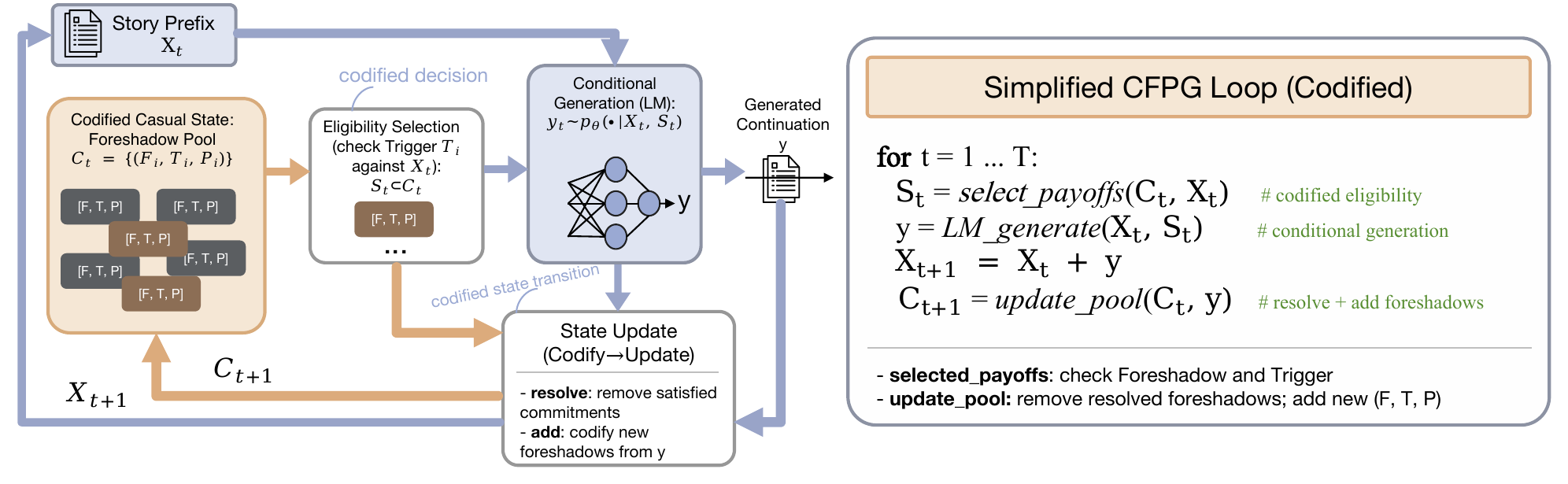}
     \caption{Overview of the CFPG framework. CFPG maintains a codified causal state in the form of a foreshadow pool $C_t$, where each element is a structured $(F, T, P)$ triple. At each step t, an eligibility selection module deterministically selects a subset $S_t \subseteq C_t$ based on codified trigger constraints, which conditions the language model to generate the next continuation $y$. The generated text updates both the narrative prefix $X_{t+1}$ and the foreshadow pool $C_{t+1}$ via a codified state transition that resolves satisfied commitments and introduces new foreshadows. The right panel shows the simplified CFPG loop in pseudocode.
}
    \label{fig:cfpg}
\end{figure*}

Despite recent advances in story generation~\citep{openai2025gpt5, googlegemini}, language models frequently fail to correctly realize such foreshadowed commitments. While generated text may remain grammatically fluent and locally coherent, models often neglect previously introduced setups, contradict established conditions, or resolve them inappropriately~\citep{see2019massively, guan2021long}. Existing story-generation benchmarks, such as ROCStories~\citep{mostafazadeh2016corpus} and WritingPrompts~\citep{fan-etal-2018-hierarchical}, primarily emphasize short-range coherence, evaluating whether each sentence follows naturally from the preceding context~\citep{rashkin2020plotmachines}. They do not explicitly test whether narrative setups are later paid off, leaving a core aspect of narrative competence underexamined~\citep{sun2021long}.

This limitation is not addressed by prior work in open-ended story generation and creative writing systems. Many approaches focus on maintaining character consistency~\citep{shanahan2023role}, dialogue coherence~\citep{shuster2022blenderbot}, or stylistic alignment~\citep{liu2023context}, often through persona conditioning~\citep{peng2025codifying} or memory-based mechanisms~\citep{wang2025generating}. Crucially, these methods primarily model persistent attributes, rather than discrete narrative events whose realization can be causally verified. While such techniques help preserve surface-level continuity, they do not provide explicit mechanisms for representing, tracking, or verifying whether foreshadowed narrative conditions are ultimately fulfilled~\citep{yang2023doc}. As a result, models may appear coherent while silently violating narrative commitments introduced earlier in the story.

To address this gap, we introduce \textbf{Codified Foreshadowing–Payoff Generation (CFPG)}, a framework that reformulates narrative coherence in terms of explicit causal realization. Rather than treating continuity as an implicit property of text, CFPG represents each foreshadow as a structured predicate specifying a \emph{foreshadow}, a \emph{trigger}, and an expected \emph{payoff}. This representation allows narrative generation and evaluation to be grounded in whether each introduced commitment is realized, postponed, or violated as the story unfolds.

By encoding foreshadow–payoff relations as executable symbolic rules, CFPG enables precise, temporally grounded detection of payoff realization that cannot be reliably achieved through prompting or context augmentation techniques alone. Moreover, this formulation provides a controllable interface for evaluating language models on narrative competence defined by commitment fulfillment, rather than surface-level fluency.

Using the \textsc{BookSum} corpus of long-form literary summaries, we automatically mine structured interpretable foreshadow–payoff pairs. This dataset supports systematic analysis of narrative commitment realization across diverse genres and story structures.

Our contributions are threefold:
\begin{enumerate}[itemsep=0pt, topsep=2pt, parsep=0pt]
    \item We introduce a novel framework that explicitly models narrative coherence through foreshadow–payoff realization, formalizing storytelling competence as commitment fulfillment.
    \item We construct a large-scale dataset derived from \textsc{BookSum}, automatically extracting explicit foreshadow–payoff pairs from long-form narratives aligned with their narrative positions.
    \item Through controlled experiments, we demonstrate that codifying narrative commitments as Foreshadow–Trigger–Payoff predicates substantially improves payoff realization accuracy and narrative alignment compared to standard prompting baselines.
\end{enumerate}

%% file: 2-rel.tex
\section{Related Works}
\subsection{Creative Writing}
Prior work on story writing with language models has primarily focused on generating fluent and coherent narratives, often emphasizing local plausibility and stylistic consistency~\citep{mostafazadeh2016corpus, fan-etal-2018-hierarchical}. As a result, models can produce locally coherent text while neglecting or contradicting previously introduced narrative elements.
To improve global coherence, several approaches incorporate planning mechanisms, guiding generation with high-level plot structures or event sequences~\citep{riedl2010narrative,yang2023doc,wang2025generating}. While these methods improve structural consistency, they do not explicitly represent narrative commitments: foreshadowed events remain implicit, and payoff realization is neither enforced nor causally verifiable.
More recent LLM-based systems employ multi-agent collaboration or feedback-based refinement to steer story development~\citep{patel2024swag,bae2024collective,venkatraman2025collabstory}. These frameworks enhance creativity and engagement, but narrative progression is still governed by qualitative feedback rather than explicit causal conditions linking setups to outcomes.

In contrast, our work directly targets this gap by codifying foreshadow–payoff relations as explicit, executable narrative constraints, enabling grounded detection and controlled realization of long-range narrative commitments.

\subsection{Controllable Text Generation}
Controllable Text Generation (CTG) studies how to steer language models to satisfy explicit control conditions during generation. Early work formalized control through conditional language models and control codes, enabling generation conditioned on topics, styles, or domains~\citep{hu2017toward}. Subsequent approaches extended this idea via disentangled latent variables~\citep{keskar2019ctrl}, decoding-time guidance~\citep{dathathri2019plug}, and classifier- or energy-based interventions~\citep{holtzman2018learning,krause2021gedi}. Recent works unify these methods under content-level and attribute-level control, covering dimensions such as sentiment, topic, style, and safety~\citep{zhou2023instruction, lambert2024tulu, liang2024controllable, yun2025ultrabench}.

Despite their success, most CTG formulations treat control signals as static, non-eventive attributes applied uniformly across generation. They regulate how text is expressed, but do not model when narrative commitments introduced earlier should be resolved. As a result, models may satisfy all declared control conditions while still violating long-range narrative logic.

%% file: 3-method.tex
\section{Method}

\subsection{Structured Representation of Narrative Commitments}

To transform intuitive narrative structures into machine-actionable logic, we formalize the relationship between foreshadowing and its resolution as an explicit system of causal commitments. We argue that a complete narrative dependency consists not just of an initial setup and a final resolution, but of a specific logical "gate" that governs its timing. Consequently, we propose representing each commitment as a structured \textbf{Foreshadow--Trigger--Payoff (F--T--P)} triple:

\begin{itemize} [itemsep=0pt, topsep=2pt, parsep=0pt]
    \item \textbf{Foreshadow ($F$)}: The initial setup or narrative anomaly that establishes a "causal debt," implying that a future explanation or resolution is required.
    \item \textbf{Trigger ($T$)}: The specific narrative condition or prerequisite event that must occur for the latent foreshadow to become actionable.
    \item \textbf{Payoff ($P$)}: The concluding event that logically fulfills and resolves the commitment introduced by $F$ and activated by $T$.
\end{itemize}

This F--T--P decomposition is essential for modeling \emph{temporal appropriateness}. In \emph{The Hound of the Baskervilles} (~\autoref{fig:problem_formulation}), the missing boot ($F$) creates suspense, but remains dormant until the revelation of Stapleton's tracking method ($T$). Only then is the explanation of the boot's purpose ($P$) narratively justified. By explicitly modeling the \textbf{Trigger}, our framework distinguishes between \emph{premature payoff} (spoiling suspense) and \emph{missing payoff} (logical inconsistency).

\subsection{Codified Foreshadow-Payoff Generation}
The CFPG framework externalizes narrative causality from the model's implicit attention weights into a \textbf{codified finite-state abstraction}. This allows for symbolic tracking of narrative debts with a level of precision that pure prompting cannot achieve.

\subsubsection{Causal State and Foreshadow Pool}
Throughout the generation process, CFPG maintains a dynamic \textbf{Foreshadow Pool} $\mathcal{C}=\left\{\left(F_i, T_i, P_i\right)\right\}_{i=1}^n$. This pool serves as a global state representing all unfulfilled narrative commitments. Unlike standard autoregressive models that compress history into a hidden vector, CFPG represents the narrative state as an explicit set of resolvable predicates.

\input{figures/example.tex}
\subsubsection{The Codified Loop: Select, Generate, and Update} As illustrated in ~\autoref{fig:cfpg}, CFPG operates through an iterative \emph{Select--Generate--Update} cycle:
\paragraph{Eligibility Selection via Codification} At each step $t$, for each pending commitment in the foreshadow pool $\mathcal{C}_t$, CFPG invokes the \texttt{codify} function. This function acts as a logical gate that evaluates the current narrative context $X_t$ against the trigger $T$. Only those foreshadows whose triggers are satisfied are promoted to the active subset $S_t = \{f \in \mathcal{C}_t \mid \operatorname{codify}(X_t, f) = \text{True}\}$.

\paragraph{Guided Continuation} Given the eligible subset $S_t$, the language model is tasked with generating the next scene $y$. Unlike vanilla generation, CFPG injects the corresponding payoffs associated with $S_t$ into the model's context as explicit narrative requirements. The generation is thus formulated as:
\[
y \sim p_\theta\left(y \mid X_t, S_t\right)
\]
This ensures that the transition to the next scene is not merely a probabilistic continuation, but a guided fulfillment of active narrative debts under explicit narrative constraints.

\paragraph{State Transition and Grounding} Following generation, CFPG performs a state update. A verification module identifies which commitments from $S_t$ were successfully realized in the text $y$ and removes them from $\mathcal{C}_t$. Simultaneously, it extracts new narrative setups introduced in $y$ and encodes them as new F--T--P triples for $\mathcal{C}_{t+1}$. This ensures the causal state representation remains temporally grounded as the story unfolds.

%% file: figures/example.tex
\begin{figure*}[t]
    \centering
    \scriptsize
    \tcbset{
      enhanced,
      boxrule=0.6pt, 
      arc=1mm,
      left=1.5mm, right=1.5mm, top=0.5mm, bottom=0.5mm, 
      fonttitle=\bfseries\small,
      coltitle=white
    }
    
    \begin{minipage}[t]{0.72\linewidth}
        \begin{tcolorbox}[title=Story Context (Excerpt), colback=customBlue!5, colframe=customBlue, colbacktitle=customBlue]
            In \emph{The Hound of the Baskervilles}, Sir Henry Baskerville arrives in London to inherit the estate. Shortly afterward, a boot mysteriously disappears from his hotel room, followed by a second. No explanation is provided at this stage, leaving the incident as a strange, unsettling anomaly in the narrative flow.
        \end{tcolorbox}
    \end{minipage}\hfill
    \begin{minipage}[t]{0.26\linewidth}
        \begin{tcolorbox}[title=Foreshadow Metadata, colback=customPurple!5, colframe=customPurple, colbacktitle=customPurple]
            \textbf{Status:} Unresolved \\
            \textbf{Type:} Physical Object \\
            \textbf{Gap:} Long-range delay \\
            \textbf{Logic:} Non-trivial link
        \end{tcolorbox}
    \end{minipage}

    \vspace{1mm} 

    \begin{tcolorbox}[
      title=Extracted Foreshadow--Trigger--Payoff Triple,
      colback=customGreen!5, 
      colframe=customGreen, 
      colbacktitle=customGreen
    ]
        \begin{tabular}{@{}ll}
            \textbf{Foreshadow (F):} & One of Sir Henry’s boots goes missing without any immediate explanation or functional cause. \\
            \textbf{Trigger (T):}     & The investigation reveals the existence of the hound and Stapleton's need for a scent-based tracking tool. \\
            \textbf{Payoff (P):}      & It is revealed that the boot was stolen to train the hound specifically to hunt Sir Henry by his scent.
        \end{tabular}
    \end{tcolorbox}
    
    \caption{\textbf{A representative F-T-P triple extracted from the \textsc{BookSum} corpus.} We anchor each element to specific narrative segments, ensuring that the latent causal link is verifiable and the payoff is temporally justified.}
    \label{fig:foreshadow_example}
\end{figure*}

%% file: 4-setting.tex
\section{Dataset}

Extracting foreshadow-payoff dependencies from full-length novels at scale presents a significant challenge due to narrative noise and extreme context length. We address this by leveraging \textsc{BookSum} \citep{kryscinski2021booksum}, a corpus of human-written, hierarchical abstractive summaries that distill long-range plot points into discourse-salient events.
We construct a sentence-level foreshadow--payoff dataset through a three-stage pipeline designed to identify, verify, and filter long-range narrative dependencies with minimal thematic noise.

Given a narrative text segmented into sentences $X = (s_1, \dots, s_T)$, our objective is to recover pairs $(s_{t_f}, s_{t_p})$ such that $s_{t_f}$ introduces a concrete narrative condition and $s_{t_p}$ later resolves it through an explicit event, decision, or revelation.

\paragraph{Stage 1: Sentence-Level Candidate Identification}
We use GPT-4.1 ~\citep{openai2025gpt41} to scan abstractive summaries and extract candidate foreshadow--payoff pairs.
Each candidate is required to be anchored to two specific sentences, yielding provisional indices $(t_f, t_p)$.
This stage prioritizes recall and admits weak or noisy candidates.

\paragraph{Stage 2: Payoff Alignment Verification}
To eliminate thematic echoes and non-causal associations, we apply a symbolic verification gate to filter these candidates.
A verifier model assesses whether the narrative context window centered at the proposed resolution point $s_{t_p}$ constitutes a genuine causal or narrative resolution of the setup context window centered at $s_{t_f}$, rejecting pairs whose linkage is metaphorical, anticipatory, or unsupported by observable textual evidence.

\paragraph{Stage 3: Rubric-Based Filtering}
Pairs that pass Stage 2 are further subjected to a rubric-based filtering stage designed to enforce strict foreshadow validity.
Two independent verifier models evaluate each candidate pair along four dimensions:

(i) \textbf{Setup Validity}: the setup introduces a concrete narrative element (e.g., object, action, rule, or decision) that is not fully explained or resolved at the time of its introduction;

(ii) \textbf{Payoff Validity}: the payoff provides new narrative information that fulfills, resolves, or retroactively reinterprets the setup, rather than restating or trivially extending it;

(iii) \textbf{Temporal Separation}: the setup and payoff occur in distinct sentences and are separated by a non-trivial narrative interval, excluding immediate or locally resolved cause--effect relations;

(iv) \textbf{Foreshadow Justification}: the setup can be reasonably interpreted as a deliberate narrative foreshadow only in hindsight, after observing the payoff, and would otherwise remain narratively under-specified.

A foreshadow--payoff pair is retained in the final dataset only if both verifier models accept the pair on all four criteria.

\vspace{-1em}
\paragraph{Resulting Dataset}
The resulting dataset comprises sentence-anchored foreshadow--payoff instances extracted from complete narrative summaries.
For each foreshadow, we provide: (i) the full summary text, (ii) a sentence-level index $t_f$ marking the introduction of the foreshadowed setup, (iii) a sentence-level index $t_p$ identifying its verified payoff resolution, (iv) a concise natural-language description of the foreshadow--payoff relation, and (v) a categorical foreshadow type (e.g., \texttt{object}, \texttt{event}, \texttt{rule}). ~\autoref{fig:foreshadow_example} illustrates an example extracted using our pipeline. Additional statistics are provided in ~\autoref{appendix:dataset}.

%% file: 5-exp.tex
\section{Experiments}
We ask the following research questions to evaluate the effects of CFPG on narrative reasoning:
\begin{enumerate} [itemsep=0pt, topsep=2pt, parsep=0pt]
    \item \textbf{Payoff Activation under Oracle Timing}: Can CFPG reliably activate and realize eligible foreshadows when the payoff timing is externally specified? (\S 5.1)
    
    \item \textbf{Grounded Payoff Decision}: 
    Does CFPG enable grounded detection and realization of foreshadowed payoffs under incrementally revealed narrative context? (\S 5.2) 
    
    \item \textbf{Error Attribution}: 
    Where do grounded payoff decisions fail, and how does CFPG change the underlying error patterns during narrative progression? (\S 5.3)
\end{enumerate}

\subsection{Conditional Payoff Activation in Truncated Scenes}
\label{sec:payoff_activation_under_oracle_timing}

\input{tables/payoff_under_oracle_timing.tex}
We first investigate a fundamental question: 
\textit{Can CFPG reliably activate and realize eligible foreshadows when the payoff timing is externally specified?} 
Standard prompting often produces locally fluent continuations that nevertheless fail to "settle" existing narrative debts. To diagnose this, we construct a controlled experiment by truncating narratives immediately before a known payoff point (a.k.a. Oracle Timing), isolating the model's ability to transition from setup to resolution.

\paragraph{Behavioral Alignment via Narrative Entailment}
We evaluate model continuations using a three-class narrative entailment scheme \citep{peng2024quantifying}. Given a truncated scene, a generated sentence $\hat{y}$, and the gold payoff $y$, an LLM judge determines if $\hat{y}$ \textbf{entails} (1.0), is \textbf{neutral} to (0.5), or \textbf{contradicts} (0.0) the intended narrative trajectory.

\input{tables/grounded_payoff_tracking.tex}

As shown in ~\autoref{tab:payoff_under_oracle_timing}, standard prompting frequently results in payoff-agnostic continuations—plausible sentences that ignore the pending foreshadow. In contrast, CFPG achieves near-ceiling \textit{Should-Payoff Rates} ($>0.96$) and significantly higher alignment scores. By utilizing the Trigger state, CFPG guides the model to actively resolve previously introduced narrative debts, ensuring the continuation is not just fluent but logically grounded in the specific foreshadowing setup.
\vspace{-1em}
\paragraph{Mechanistic Evidence: Causal Saliency Tracking} 
To understand \textit{why} CFPG outperforms standard prompting, we analyze the model's internal attention patterns. We compare the attention weights assigned to the \textit{Setup} tokens during the generation of the payoff sentence.

\begin{figure}[htbp] 
    \centering
     \includegraphics[width=0.50\textwidth]{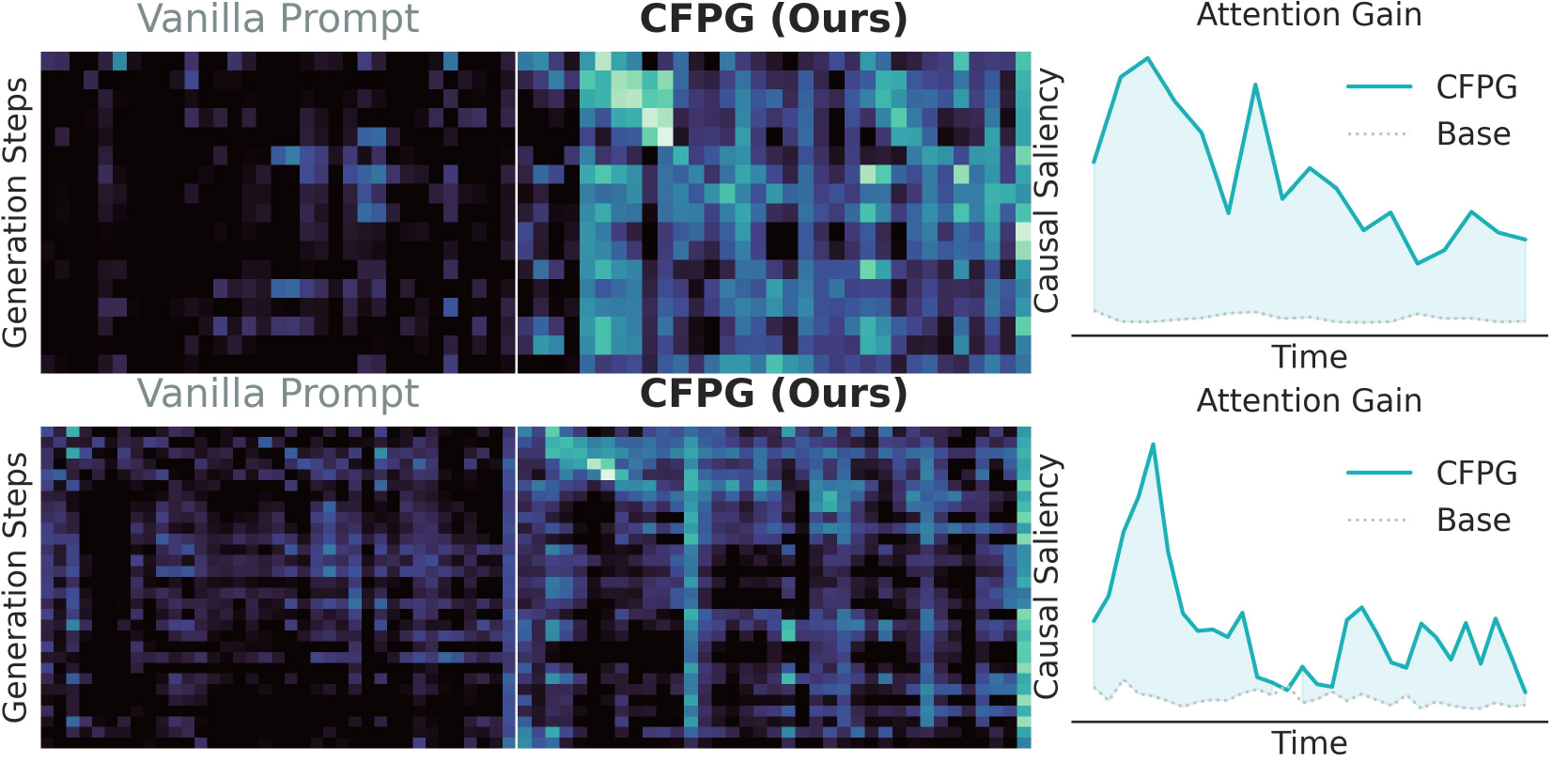}
     \caption{\textbf{Visualization of attention patterns during payoff generation.} The heatmaps (left and center) compare the attention weights allocated to foreshadowing setup tokens under the vanilla prompting baseline and CFPG. The line plot (right) quantifies the resulting Causal Saliency Gain relative to the baseline, showing that CFPG consistently maintains significantly higher mean attention to the setup region throughout the generation process. Each row corresponds to a different narrative instance.}     
    \label{fig:attn}
\end{figure}

As visualized in ~\autoref{fig:attn}, we observe a stark shift in causal saliency. In the baseline (Left), attention to the distant setup is diffuse and sparse. This suggests that the model may fail to recognize the necessity of foreshadow resolution, instead prioritizing local context to maintain surface-level fluency at the expense of long-range narrative consistency. Conversely, CFPG (Right) exhibits a dense "attention spike" focused precisely on the foreshadowing anchors. The mean attention weight to the setup region exhibits a pronounced surge under CFPG compared to the baseline, confirming that our state-coded guidance effectively "re-ground" the model in the narrative's past. 

Combined, these results demonstrate that explicitly separating payoff eligibility from surface generation enables reliable, model-agnostic control over foreshadow resolution, ensuring that narrative "hooks" are not merely introduced, but meaningfully settled.

\subsection{Grounded Payoff Tracking: From Sensing to Action}
\label{sec:grounded_payoff_tracking}
We evaluate whether CFPG facilitates \emph{grounded} identification of narrative resolution by framing payoff tracking as an online sensing problem. This setup simulates an incremental reading process where the model is exposed to the narrative sentence-by-sentence. At each step, the model must perform "causal sensing", deciding whether the observed prefix has satisfied the trigger conditions for a payoff without accessing future tokens.
To provide a granular diagnosis of model behavior, we analyze three complementary dimensions of causal awareness: 
\begin{enumerate} [itemsep=0pt, topsep=2pt, parsep=0pt]
\item \textbf{Activation Timing}: The ability to suppress "causal hallucinations" (premature triggers) until the causal conditions are satisfied; 
\item \textbf{Localization Accuracy}: The precision in pinning down the specific narrative anchor where resolution occurs; 
\item \textbf{Generative Fidelity}: The capacity to translate a successful detection into a trajectory-consistent continuation. \end{enumerate}

\paragraph{Baselines}
We compare CFPG against two prompting-based baselines that vary in how narrative foreshadow information is incorporated.
(1) \textbf{Foreshadow-Aware Prompting (FAP)} augments standard prompting by explicitly providing unresolved narrative commitments in natural language, but without enforcing executable trigger conditions or maintaining a persistent narrative state.
(2) \textbf{Foreshadow-Similarity Context Refresh (FSCR)} dynamically augments a
sliding context window with earlier sentences selected by lexical similarity
to the foreshadow, without explicit state tracking or trigger enforcement.

\paragraph{Timing Precision and Causal Gating}
First, we examine whether the model can distinguish between genuine resolution and mere semantic proximity. As summarized in ~\autoref{tab:grounded_payoff_tracking}, prompting methods frequently suffer from causal hallucinations, triggering premature payoffs (235 early triggers in GPT-4.1-mini) when a relevant character or keyword is mentioned without satisfying the logical prerequisites. CFPG acts as a logical gatekeeper, reducing these early triggers by 29.3\%. This suppression of false positives provides evidence that CFPG enables the model to "wait" for the causal trigger rather than impulsively guessing based on local heuristics.

\paragraph{The Sensing-Acting Gap}
Next, we probe the fidelity of the resolution once a payoff is detected. A critical finding is the persistent gap in baseline models between detection and realization: even when the baseline correctly identifies the payoff window, its \textit{Continuation Score} remains disproportionately low (0.453 for GPT-4.1-mini). This decoupling suggests that passive recognition does not imply active commitment. In contrast, CFPG bridges this gap by anchoring the generation in a codified F--T--P state, yielding a 43\% improvement in trajectory alignment. This confirms that the explicit state serves as a functional bridge, translating a successful "sensing" event into a "consistent action."

\paragraph{Decision Dynamics of Grounded Payoff Detection}
\begin{figure}[htbp] 
    \centering
     \includegraphics[width=0.49\textwidth]{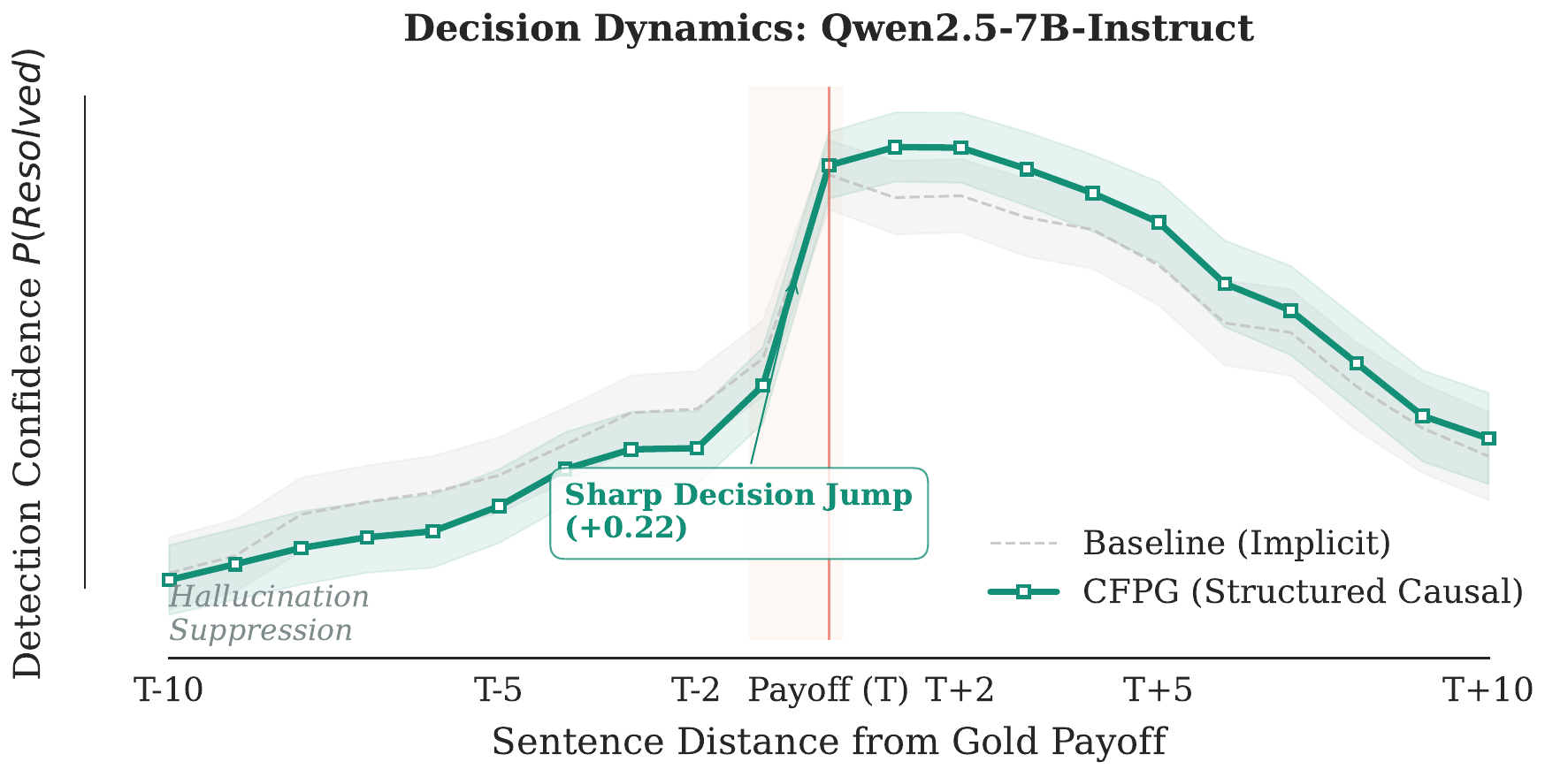}
     \caption{Temporal decision dynamics of payoff detection for Qwen-2.5-7B-Instruct. CFPG shows reduced premature activation, a sharp decision transition at the gold payoff, and sustained post-resolution confidence compared to the baseline.}     
    \label{fig:dynamics_analysis}
\end{figure}
To further understand \emph{why} CFPG improves grounded payoff tracking beyond static accuracy gains, we analyze the \emph{decision dynamics} of payoff detection under strictly incremental observation settings. Instead of treating detection as a point estimate, we probe how the model’s internal confidence evolves as narrative context approaches the gold payoff location.

Concretely, we measure the model’s activation confidence for the binary decision \emph{``Should the payoff occur now?''} at successive sentence prefixes centered around the gold payoff index. This yields a temporal confidence trajectory that reveals whether payoff recognition emerges as a discrete causal decision or as a gradual semantic drift.

~\autoref{fig:dynamics_analysis} reveals a clear contrast in how FAP prompting and CFPG distribute payoff activation over narrative time. Prior to the true payoff point, the baseline exhibits a consistently higher activation probability. This elevated pre-payoff confidence reflects premature causal commitment rather than grounded recognition.

Around the gold payoff boundary, both methods show a sharp increase in activation; however, CFPG displays a markedly steeper and more localized decision jump (+0.22), suggesting a discrete transition from “unresolved” to “resolved” once the causal prerequisites are met. Crucially, after the payoff point, CFPG sustains a high activation level over an extended window, whereas the baseline rapidly decays. This post-payoff persistence indicates that CFPG maintains a stable causal state once resolution is achieved.

Together, these dynamics suggest that CFPG does not merely improve payoff timing accuracy, but fundamentally reshapes the decision process: suppressing premature activation before the payoff, enforcing a sharp causal switch at resolution, and preserving commitment afterward.

\subsection{Error Attribution in Grounded Payoff Tracking}
\label{sec:error_attribution}

\input{tables/error_types}
To better understand the failure modes of grounded payoff tracking, we conduct a systematic error attribution analysis over all incorrect end-to-end predictions.
This analysis focuses on \emph{decision-level behavior} exhibited by the model during incremental narrative processing—specifically, how payoff-related decisions are made, delayed, or prematurely triggered given partial context.

\paragraph{Analysis Procedure.}
For each failure case, we first elicit a concise, text-grounded rationale explaining why the model made an incorrect payoff decision under the available context. Explanations are generated using a constrained prompt that enforces explicit reference to observable narrative evidence and disallows speculation about future events or alternative plot developments. This ensures that all rationales reflect concrete grounding failures rather than post-hoc interpretation.

We then induce an error taxonomy in a data-driven manner by clustering these rationales into a small set of recurring failure patterns. This process yields a compact taxonomy that captures distinct mechanisms of grounded payoff failure without relying on predefined labels. Finally, each failure instance is assigned to its dominant category, enabling aggregate analysis of error distributions across methods.

\vspace{-1em}
\paragraph{Error Taxonomy.}
The induced taxonomy consists of six recurring error types (\autoref{tab:error_taxonomy}):
(1) \emph{Premature Payoff Triggering}, where payoff decisions are made based on surface cues or anticipatory signals before concrete realization;
(2) \emph{Deferred Payoff Not Yet Observable}, where the model loses track of the pending status and prematurely assumes resolution during a long narrative gap;
(3) \emph{Thematic or Event-Level Confusion}, where related but distinct events or motifs are mistaken for the true payoff;
(4) \emph{Narrative State Tracking Failure}, where causal commitments are not consistently maintained or updated over time;
(5) \emph{Overly Conservative Triggering}, where payoff detection is delayed due to excessive evidentiary requirements;
and (6) \emph{Indirect or Retrospective Payoff Linking Failure}, where payoffs are expressed implicitly or retrospectively but fail to be linked back to the original setup.

\begin{figure}[htbp] 
    \centering
     \includegraphics[width=0.495\textwidth]{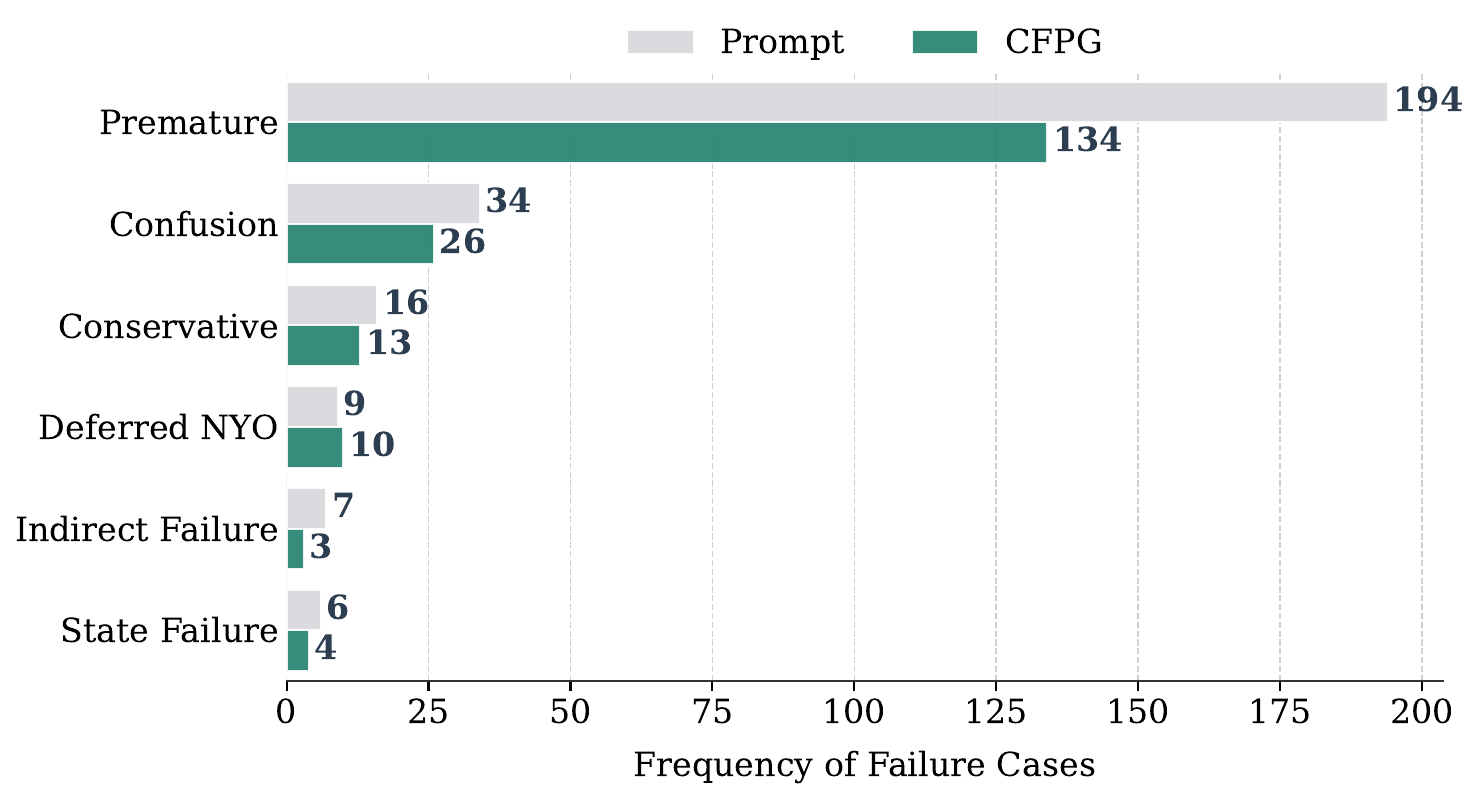}
     \caption{Distribution of grounded payoff tracking errors for prompt-based and CFPG-based methods. CFPG notably attenuates premature payoff triggering compared to baseline prompting.
     }        
    \label{fig:error_attribution}
\end{figure}

\paragraph{Findings.}
As shown in ~\autoref{fig:error_attribution}, the distribution of failure cases highlights that Premature triggering is the dominant error mode for both models. CFPG significantly outperforms the Prompt baseline by reducing Premature cases by 31\%, demonstrating its robust ability to filter semantic hallucinations.

Regarding high-level narrative reasoning, CFPG exhibits a marked improvement in maintaining \textbf{logical consistency} and capturing \textbf{complex associations}. Specifically, the reduction in \textit{Thematic Confusion} (from 34 to 26 cases) and the near-halving of \textit{Indirect Failure} cases (7 to 3)  suggest that the model successfully links implicit or non-linear payoffs to their original setups even across intricate narrative structures.

%% file: tables/payoff_under_oracle_timing.tex

\begin{table}[t]
    \centering
    \small
    \caption{\textbf{Performance comparison on the BookSum dataset under oracle timing.} CFPG consistently outperforms prompt-based baselines across multiple model architectures, achieving near-perfect payoff activation and significantly improved narrative alignment.}
    \label{tab:payoff_under_oracle_timing}
    \resizebox{.49\textwidth}{!}{
    \begin{tabular}{@{} l l c S[table-format=1.3] @{}} 
        \toprule
        \textbf{Base Model} & \textbf{Method} & \textbf{Should-Payoff Rate} & \textbf{Avg. Score $\uparrow$} \\
        \midrule
        \multirow{2}{*}{GPT-4.1-mini} & Prompt & --- & 0.569 \\
         & \cellcolor{gray!10}CFPG & \cellcolor{gray!10}1.000 & \cellcolor{gray!10}\textbf{0.911} \\
        \addlinespace[0.5em]
        
        \multirow{2}{*}{Claude-Haiku-4.5} & Prompt & --- & 0.657 \\
         & \cellcolor{gray!10}CFPG & \cellcolor{gray!10}0.965 & \cellcolor{gray!10}\textbf{0.940} \\
        \addlinespace[0.8em] 
        
        \multirow{2}{*}{Qwen2.5-3B} & Prompt & --- & 0.481 \\
         & \cellcolor{gray!10}CFPG & \cellcolor{gray!10}0.998 & \cellcolor{gray!10}\textbf{0.781 }\\
        \addlinespace[0.3em]
        
        \multirow{2}{*}{Qwen2.5-7B} & Prompt & --- & 0.517 \\
         & \cellcolor{gray!10}CFPG & \cellcolor{gray!10}1.000 & \cellcolor{gray!10}\textbf{0.797} \\
        \addlinespace[0.3em]
        
        \multirow{2}{*}{Qwen2.5-14B} & Prompt & --- & 0.583 \\
         & \cellcolor{gray!10}CFPG & \cellcolor{gray!10}1.000 & \cellcolor{gray!10}\textbf{0.898 }\\
        \addlinespace[0.8em]
        
        \multirow{2}{*}{Llama-3.1-8B} & Prompt & --- & 0.530 \\
         & \cellcolor{gray!10}CFPG & \cellcolor{gray!10}1.000 & \cellcolor{gray!10}\textbf{0.802} \\
        \bottomrule
        \addlinespace[0.2em]
        \multicolumn{4}{@{}l}{\footnotesize \textit{Note: All models refer to their respective ``Instruct'' variants.}} \\
    \end{tabular}
    }
\end{table}

%% file: tables/grounded_payoff_tracking.tex
\begin{table*}[t]
    \centering
    \small
    \caption{\textbf{Main Results: Grounded Payoff Tracking.} We evaluate the model's ability to identify payoffs under incremental context. \textbf{CFPG (Ours)} consistently achieves superior performance across all backbones.}
    \label{tab:grounded_payoff_tracking}
    
    \definecolor{lightgray}{gray}{0.96}
    \newcommand{\gain}[1]{\text{\scriptsize\textcolor{teal}{+#1}}}
    \newcommand{\loss}[1]{\text{\scriptsize\textcolor{teal}{-#1}}}
    \newcommand{\bad}[1]{\text{\scriptsize\textcolor{red}{+#1}}}

    \setlength{\tabcolsep}{5pt} 
    \renewcommand{\arraystretch}{1.2}
    
    \resizebox{.9\textwidth}{!}{
        \begin{tabular}{l l c c c c c}
            \toprule
            & & \textbf{Activation} & \multicolumn{3}{c}{\textbf{Localization (Online Sensing)}} & \textbf{Generation} \\
            \cmidrule(lr){3-3} \cmidrule(lr){4-6} \cmidrule(l){7-7}
            \textbf{Model} & \textbf{Method} & \textbf{Det. (\%) $\uparrow$} & \textbf{Early $\downarrow$} & \textbf{Late $\downarrow$} & \textbf{Error $\downarrow$} & \textbf{Fidelity $\uparrow$} \\
            \midrule
            
            \multirow{3}{*}{GPT-4.1-mini} 
                & Foreshadow-Aware Prompt & 58.0 & 235 & 17 & 8.85 & 0.453 \\
                & Context Refresh (Sim.) & 48.6 & 306 & 7 & 13.05 & 0.382 \\
                & \cellcolor{gray!10} \textbf{CFPG (Ours)} & \cellcolor{gray!10}\textbf{69.8} \gain{11.8} & \cellcolor{gray!10}\textbf{166} \loss{69} & \cellcolor{gray!10}\textbf{11} \loss{6} & \cellcolor{gray!10}\textbf{5.76} \loss{3.09} & \cellcolor{gray!10}\textbf{0.647} \gain{0.221} \\
            
            \midrule
            \multirow{3}{*}{\shortstack[l]{Qwen2.5\\3B-Inst.}} 
                & Foreshadow-Aware Prompt & 4.5 & 601 & 0 & 34.67 & 0.022 \\
                & Context Refresh (Sim.) & 6.7 & 587 & 0 & 32.81 & 0.039 \\
                & \cellcolor{gray!10} \textbf{CFPG (Ours)} & \cellcolor{gray!10}\textbf{10.8} \gain{6.3} & \cellcolor{gray!10}\textbf{550} \loss{51} & \cellcolor{gray!10}5 \bad{5} & \cellcolor{gray!10}\textbf{31.63} \loss{3.04} & \cellcolor{gray!10}\textbf{0.080} \gain{0.058} \\
            
            \midrule
            \multirow{3}{*}{\shortstack[l]{Qwen2.5\\7B-Inst.}} 
                & Foreshadow-Aware Prompt & 15.6 & 522 & 2 & 27.00 & 0.114 \\
                & Context Refresh (Sim.) & 19.6 & 493 & 7 & \textbf{23.79} & 0.177 \\
                & \cellcolor{gray!10} \textbf{CFPG (Ours)} & \cellcolor{gray!10} 19.6 \gain{4.0} & \cellcolor{gray!10}\textbf{490} \loss{32} & \cellcolor{gray!10}\textbf{5} \bad{3} & \cellcolor{gray!10} 26.23 \loss{0.77} & \cellcolor{gray!10}\textbf{0.184} \gain{0.070} \\
                
            \midrule
            \multirow{3}{*}{\shortstack[l]{Llama3.1\\8B-Inst.}} 
                & Foreshadow-Aware Prompt & 25.8 & 451 & 5 & 20.04 & 0.182 \\
                & Context Refresh (Sim.) & 22.1 & 482 & 3 & 23.72 & 0.160 \\
                & \cellcolor{gray!10} \textbf{CFPG (Ours)} & \cellcolor{gray!10} \textbf{27.3} \gain{1.5} & \cellcolor{gray!10} \textbf{439} \loss{12} & \cellcolor{gray!10} 8 \bad{3} & \cellcolor{gray!10} \textbf{19.91} \loss{0.13} & \cellcolor{gray!10}\textbf{0.226} \gain{0.044} \\
                
            \bottomrule
            \multicolumn{7}{@{}l}{\footnotesize \textit{Note: Definitions of all metrics are provided in~\autoref{appendix:metrics}. Colored values denote the difference between CFPG and FAP.}} \\
        \end{tabular}
    }
\end{table*}

%% file: tables/error_types.tex
\begin{table*}[htbp]
\centering
\small
\caption{Taxonomy of Grounded Payoff Tracking Failures}
\label{tab:error_taxonomy}
\begin{tabularx}{\textwidth}{@{} l l X X @{}}
\toprule
\rowcolor{groupgray}
\textbf{Category} & \textbf{Decision} & \textbf{Model's Internal Logic} & \textbf{Ground Truth (Narrative State)} \\ \midrule

\multicolumn{4}{l}{\textit{\textbf{\color{myred}Active Errors (False Triggers)}}} \\
Premature & Early & ``The intent or preparation is already enough.'' & Only setup/intent; causal chain not closed. \\
Confusion & Spurious & ``This event \textbf{looks} like the payoff point.'' & Thematic similarity; not the target event. \\ \midrule

\multicolumn{4}{l}{\textit{\textbf{\color{myblue}Passive Errors (State Breakdowns)}}} \\
Deferred NYO & Random & ``I \textbf{guess} it's resolved during this gap.'' & Causal state must remain \textit{Pending}. \\
State Failure & Disorder & ``I lost track of the initial \textbf{status} change.'' & Environment/character state has shifted. \\ \midrule

\multicolumn{4}{l}{\textit{\textbf{\color{mygreen}Reactive Errors (Detection Gaps)}}} \\
Conservative & Late & ``Too subtle; I'll wait for explicit confirmation.'' & Payoff occurred; model threshold too high. \\
Indirect Failure & Missed & ``There is \textbf{no link} between these two events.'' & Indirect or retrospective causal linking. \\ \bottomrule
\end{tabularx}
\end{table*}

%% file: 6-con.tex
\section{Conclusion}
We presented Codified Foreshadowing–Payoff Generation (CFPG), a framework that models narrative coherence as the explicit realization of causal commitments. By representing foreshadows as executable Foreshadow–Trigger–Payoff predicates and tracking them as a codified state, CFPG enables grounded payoff detection and controlled realization under incremental context. Experiments show consistent improvements over prompt-based baselines in payoff timing, localization, and narrative alignment.

%% file: 7-lim.tex
\section*{Limitations}
CFPG primarily targets explicit and textually grounded foreshadow payoff relations and does not aim to model highly abstract or purely symbolic narrative devices. Our experiments are conducted on summary level narratives, which provide a controlled setting for evaluating long range causal dependencies but may not capture all stylistic or discourse level phenomena in full length texts. In addition, CFPG relies on automatically extracted Foreshadow Trigger Payoff structures, and extraction errors or omissions may limit coverage in some cases.

%% file: appendix.tex
\section{Dataset}
\label{appendix:dataset}

\subsection{Statistics}
The resulting dataset consists of 629 validated foreshadow-payoff pairs extracted from 148 books (\autoref{tab:dataset_stats}).

\input{tables/dataset_stats}

\begin{figure}[htbp] 
    \centering
     \includegraphics[width=0.50\textwidth]{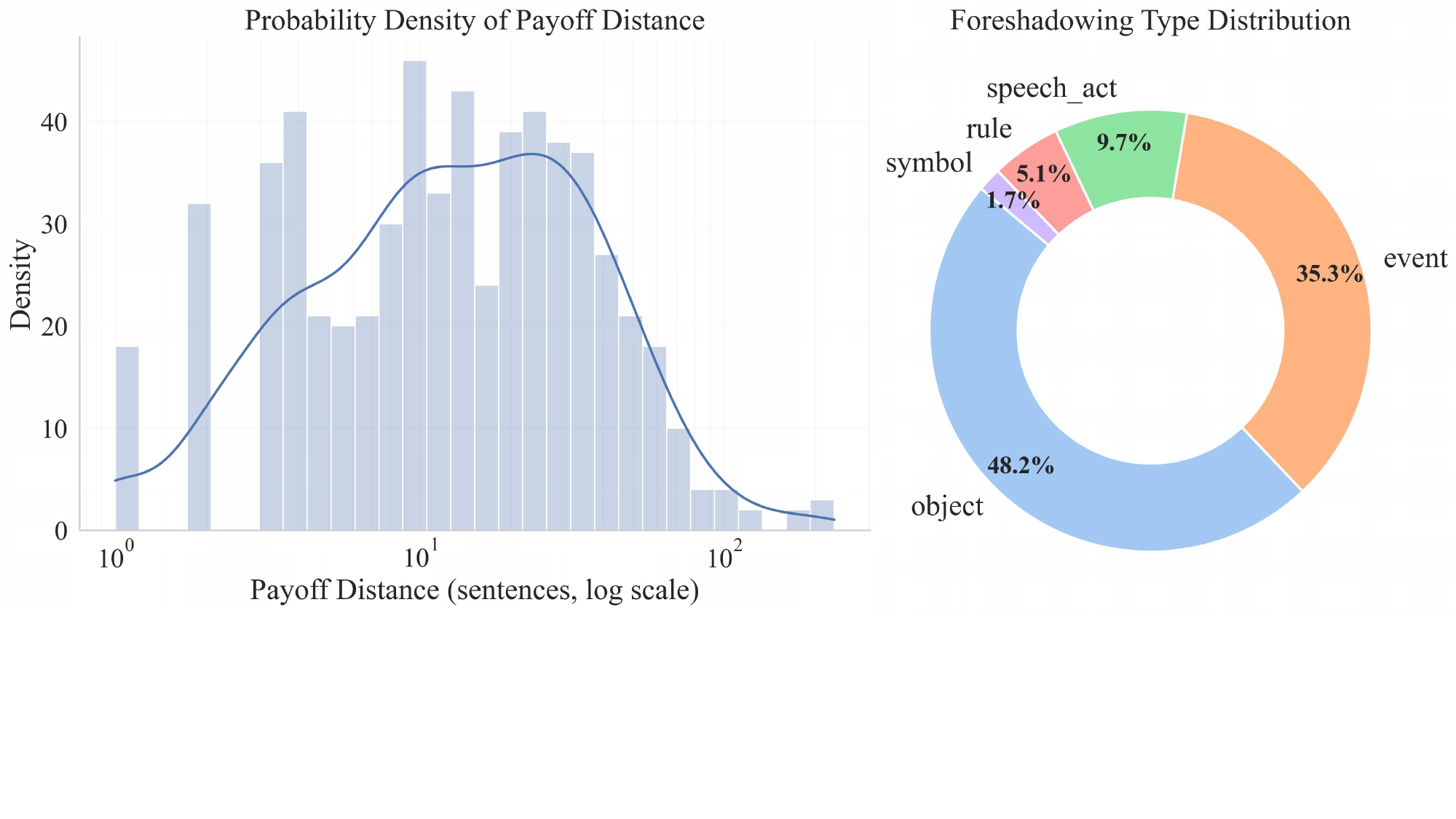}
     \caption{Dataset statistics of the extracted foreshadow--payoff corpus.}     
    \label{fig:dataset_stats}
\end{figure}

~\autoref{fig:dataset_stats} (Left) illustrates the probability density of payoff distance measured in sentences between the setup and payoff.
The distribution exhibits a pronounced heavy tail, with a median distance of 13 sentences and a mean of 20.9 sentences.
Notably, 25\% of payoffs occur at distances greater than 29 sentences, and 10\% exceed 45 sentences, with the longest dependencies spanning over 200 sentences.
This long-range structure indicates that a substantial portion of the dataset requires maintaining unresolved causal commitments across extended narrative context rather than relying on local cause--effect relations.

~\autoref{fig:dataset_stats} (Right) shows the distribution of foreshadow types.
Object-based (48.2\%) and event-based (35.3\%) foreshadows together account for more than 80\% of the dataset, reflecting the predominance of concrete narrative elements that participate in delayed causal structure.
Speech-act-based and rule-based foreshadows form a smaller but non-trivial fraction, while symbol-based foreshadows are rare (1.7\%), consistent with our conservative filtering of purely thematic or interpretive cues.

Overall, these statistics confirm that the dataset emphasizes explicit, long-range causal dependencies grounded in observable narrative events, providing a challenging and well-controlled testbed for evaluating narrative generation under delayed causal constraints.

\subsection{Quality Check and Reliability Analysis}

\input{tables/reliability}

To assess annotation reliability, we conduct a quality check on a random sample of 100 extracted foreshadow--payoff pairs.
Two annotators independently evaluate each sample from different perspectives, assessing both overall validity and component-level correctness.

As shown in ~\autoref{tab:quality_check}, the annotators agree on 88\% of the samples at the pair level.
Component-wise agreement is higher, with 95\% agreement on setup accuracy and perfect agreement on payoff accuracy.
Judgments on connection validity exhibit slightly lower agreement (88\%), reflecting the greater subjectivity involved in assessing long-range narrative reinterpretation compared to identifying explicit textual evidence.

Overall, these results indicate that the dataset achieves high precision in identifying concrete setups and payoffs, with remaining disagreements primarily arising from borderline cases of causal relevance rather than factual inconsistencies.

\section{Metric Definitions for Grounded Payoff Tracking}
\label{appendix:metrics}

We evaluate grounded payoff tracking under an \emph{incremental narrative setting},
where the model processes the story sentence by sentence and is not allowed
to revise past decisions.

\paragraph{Correct Detection Rate (Correct Det. \%).}
Correct Detection Rate measures the proportion of narratives in which the model
correctly identifies the occurrence of a payoff during incremental processing.
A prediction is considered correct if the model triggers a payoff decision at
a sentence index that lies within a fixed tolerance window ($\pm 3$ sentences) around
the annotated ground-truth payoff location.
This metric reflects end-to-end payoff detection accuracy under partial context.

\paragraph{Early Triggers.}
Early Triggers counts the number of cases in which the model predicts a payoff
\emph{before} any ground-truth payoff becomes observable in the narrative.
Such errors correspond to premature payoff triggering, typically caused by
surface-level cues or anticipatory signals that precede actual realization.
Lower values indicate better resistance to premature causal inference.

\paragraph{Late Triggers.}
Late Triggers counts the number of cases in which the model triggers a payoff
\emph{after} the ground-truth payoff point has already occurred.
These errors reflect delayed recognition or overly conservative decision-making,
where excessive evidence is required before committing to a payoff decision.
Lower values indicate better temporal alignment with the narrative payoff.

\paragraph{Localization Error (Loc. Error).}
Localization Error measures the absolute distance, in number of sentences,
between the model's predicted payoff trigger point and the annotated
ground-truth payoff location, averaged over all triggered cases.
This metric evaluates temporal precision in payoff localization during
online narrative sensing.
Lower values indicate more accurate alignment with the true payoff position.

\paragraph{Continuation Score (Cont. Score).}
Continuation Score evaluates the quality of payoff realization in generation
once a payoff has been detected.
For cases where the model triggers a payoff within the tolerance window,
we prompt the model to generate a one-sentence continuation conditioned on the
available context.
The generated continuation is then compared against the ground-truth continuation
using a trajectory-based evaluator that assesses narrative consistency,
causal progression, and outcome alignment.
The score reflects the fraction of continuations judged to follow the same
narrative trajectory as the ground truth.

\paragraph{Metric Summary.}
Together, these metrics evaluate grounded payoff tracking along three complementary
dimensions: (i) decision timing (Correct Detection, Early/Late Triggers),
(ii) temporal localization precision (Localization Error), and
(iii) generation fidelity after detection (Continuation Score),
providing a comprehensive assessment of both detection and realization
under incremental narrative context.

\section{LLM Usage Statement}
Large language models (LLMs) were employed exclusively to polish wording and improve grammatical correctness for better readability. They were not used to generate, alter, or influence the paper’s technical content, conceptual contributions, methods, or experimental findings. Responsibility for the final manuscript and its accuracy remains entirely with the authors.

%% file: tables/dataset_stats.tex
\begin{table}[htbp]
    \centering
    \small
    \caption{Dataset statistics for the extracted foreshadow--payoff pairs.}
    \begin{tabular}{l r}
    \toprule
    \textbf{Statistic} & \textbf{Value} \\
    \midrule
    \# Books & 148 \\
    \# Foreshadows & 629 \\
    \midrule
    Avg. Payoff Distance (sentences) & 20.9 \\
    Median Payoff Distance & 13.0 \\
    75th Percentile Distance & 29.0 \\
    90th Percentile Distance & 45.0 \\
    Max Payoff Distance & 230 \\
    \midrule
    Object Foreshadows & 48.2\% \\
    Event Foreshadows & 35.3\% \\
    Speech-act Foreshadows & 9.7\% \\
    Rule Foreshadows & 5.1\% \\
    Symbol Foreshadows & 1.7\% \\
    \midrule
    Avg. Extraction Confidence & 0.98 \\
    \bottomrule
    \end{tabular}
    \label{tab:dataset_stats}
\end{table}

%% file: tables/reliability.tex
\begin{table}[t]
\centering
\small
\caption{Quality check results on a random sample of 100 foreshadow--payoff pairs.
Two annotators independently evaluate overall validity and component-level correctness.
Agreement denotes the proportion of samples receiving identical judgments.}
\begin{tabular}{l c c c}
\toprule
\textbf{Aspect} & \textbf{A} & \textbf{B} & \textbf{Agree.} \\
\midrule
Pair Validity       & 0.89 & 0.96 & 0.87 \\
\midrule
Setup Accuracy      & 0.97 & 0.98 & 0.95 \\
Payoff Accuracy     & 1.00 & 1.00 & 1.00 \\
Connection Validity & 0.90 & 0.96 & 0.88 \\
\bottomrule
\end{tabular}
\label{tab:quality_check}
\end{table}